\documentclass[10pt,twocolumn,letterpaper]{article}

\usepackage{iccv}
\usepackage{times}
\usepackage{epsfig}
\usepackage{graphicx}
\usepackage{amsmath}
\usepackage{amssymb}
\newcommand{\ch}{\checkmark}
\usepackage{times}
\usepackage{epsfig}
\usepackage{graphicx}
\usepackage{amsmath,amssymb} 
\usepackage{amssymb}
\usepackage[export]{adjustbox}
\usepackage{array}
\usepackage{rotating}
\usepackage{multirow}
\usepackage{marvosym}
\usepackage{booktabs}

\usepackage[pagebackref=true,breaklinks=true,letterpaper=true,colorlinks,bookmarks=false]{hyperref}

\iccvfinalcopy 

\def\iccvPaperID{3246} 

\ificcvfinal\fi
\begin{document}

\title{Real Image Denoising with Feature Attention}
\author{Saeed Anwar~\thanks{\Letter:~saeed.anwar@csiro.au}, Nick Barnes~\thanks{\Letter:~nick.barnes@csiro.au}\\
Data61, CSIRO and The Australian National University, Australia.}

\def\NB#1{{\color{red} {{#1}}}} 
\def\SA#1{{\color{blue} {{#1}}}} %
\maketitle

\begin{abstract}
Deep convolutional neural networks perform better on images containing spatially invariant noise (synthetic noise); however, their performance is limited on real-noisy photographs and requires multiple stage network modeling. To advance the practicability of denoising algorithms, this paper proposes a novel single-stage blind real image denoising network (RIDNet) by employing a modular architecture. We use a residual on the residual structure to ease the flow of low-frequency information and apply feature attention to exploit the channel dependencies. Furthermore, the evaluation in terms of quantitative metrics and visual quality on three synthetic and four real noisy datasets against 19 state-of-the-art algorithms demonstrate the superiority of our RIDNet.
\end{abstract}

\section{Introduction}

Image denoising is a low-level vision task that is essential in a number of ways. First of all, during image acquisition, some noise corruption is inevitable and can downgrade the visual quality considerably; therefore, removing noise from the acquired image is a key step for many computer vision and image analysis applications~\cite{gonzalez1977DIP}. Secondly, denoising is a unique testing ground for evaluating image prior and optimization methods from a Bayesian perspective~\cite{Gu2014WNN,Zoran2011EPLL}. Furthermore, many image restoration tasks can be solved in the unrolled inference through variable splitting methods by a set of denoising subtasks, which further widens the applicability of image denoising~\cite{afonso2010fast,heide2014flexisp,romano2017little,zhang2017IRCNN}.

 Generally, denoising algorithms can be categorized as model-based and learning-based. Model-based algorithms include non-local self-similarity (NSS)~\cite{Dabov2007BM3D,Buades2005NLM,Dabov2009BM3DSAPCA},  sparsity~\cite{Gu2014WNN,peng2012rasl},  gradient methods~\cite{Osher2005Iterative,xu2007Iterative,weiss2007makes}, Markov random field models~\cite{roth2009fields}, and external denoising priors~\cite{anwar2017category,Yue2014CID,luo2015adaptive}. The model-based algorithms are computationally expensive, time-consuming, unable to suppress the spatially variant noise directly and characterize complex image textures. On the other hand, discriminative learning aims to model the image prior from a set of noisy and ground-truth image sets. One technique is to learn the prior in steps in the context of truncated inference~\cite{chen2017TNRD} while another approach is to employ brute force learning, for example, MLP~\cite{Burger2012MLP} and CNN methods~\cite{zhang2017DnCNN,zhang2017IRCNN}. CNN models~\cite{zhang2018ffdnet,guo2018CBDnet} improved denoising performance, due to 
 their modeling capacity, network training, and design.  However, the performance of the current learning models is limited and tailored for a specific level of noise.
 
 A practical denoising algorithm should be efficient, flexible, perform denoising using a single model and handle both spatially variant and invariant noise when the noise standard-deviation is known or unknown. Unfortunately, the current state-of-the-art algorithms are far from achieving all of these aims.
 We present a CNN model which is efficient and capable of handling synthetic as well as real-noise present in images. We summarize the contributions of this work in the following paragraphs.
 
 \begin{figure}
\begin{center}
\begin{tabular}[b]{c@{ }c@{ }c} 
      
\includegraphics[trim={5cm 2cm  5cm  2cm },clip,width=.15\textwidth,valign=t]{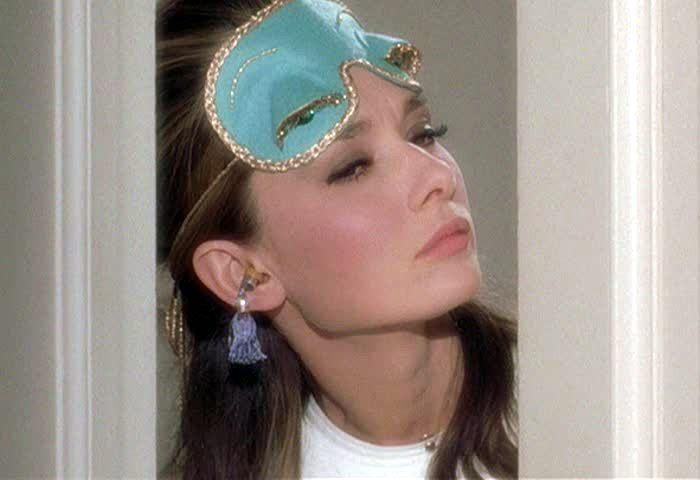}&   
\includegraphics[trim={5cm 2cm  5cm  2cm },clip,width=.15\textwidth,valign=t]{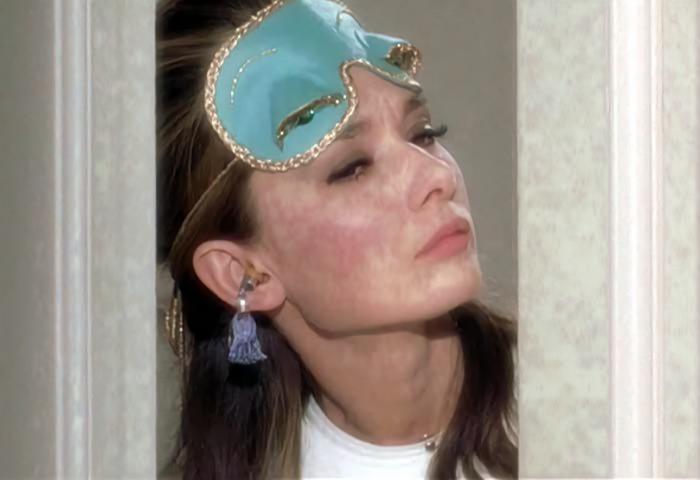}&
\includegraphics[trim={5cm 2cm  5cm  2cm },clip,width=.15\textwidth,valign=t]{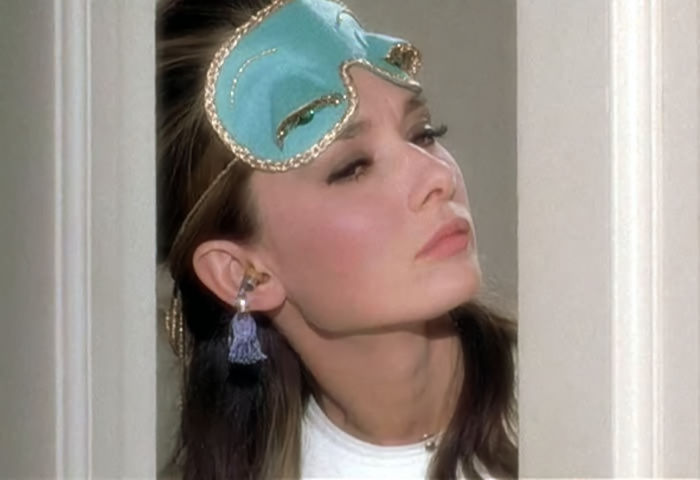}\\
Noisy & CBDNet~\cite{guo2018CBDnet} & RIDNet (Ours)\\
\end{tabular}
\end{center}
\caption{A real noisy face image from RNI15 dataset~\cite{lebrun2015NC}. Unlike 
CBDNet~\cite{guo2018CBDnet}, RIDNet does not have over-smoothing or over-contrasting artifacts (Best viewed in color on high-resolution display)}
\label{fig:Nam}
\vspace*{-5mm}
\end{figure}

\subsection{Contributions}
\begin{itemize}
\item Present CNN based approaches for real image denoising employ two-stage models; we present the first model that provides state-of-the-art results using only one stage. 

\item  
To best of our knowledge, our model is the first to incorporate feature attention in denoising.


\item Most current models connect the weight layers consecutively; and so increasing the depth will not help improve performance~\cite{dong2016SRCNN,lim2017EDSR}.
Also, such networks can suffer from vanishing gradients~\cite{bengio1994vanishing}.  We present a modular network, where increasing the number of modules helps improve performance.

\item We experiment on three synthetic image datasets and four real-image noise datasets to show that our model achieves state-of-the-art results on synthetic and real images quantitatively and qualitatively.
\end{itemize}

\section{Related Works}
In this section, we present and discuss recent trends in the image denoising.  Two notable denoising algorithms, NLM~\cite{Buades2005NLM} and BM3D~\cite{Dabov2007BM3D}, use self-similar patches.  Due to their success, many variants were proposed, including SADCT~\cite{Foi2007SADCT},  SAPCA~\cite{Dabov2009BM3DSAPCA}, NLB~\cite{Lebrun2013NLB}, and INLM~\cite{Goossens2008INLM} which seek self-similar patches in different transform domains. Dictionary-based methods~\cite{Elad2009ERD,Mairal2009NLSM,Dong2011CSR} enforce sparsity by employing self-similar patches and learning over-complete dictionaries from clean images. Many algorithms \cite{Zoran2011EPLL, Chen2015External, Xu2015PG-GMM} investigated the maximum likelihood algorithm to learn a statistical prior, \eg the Gaussian Mixture Model of natural patches or patch groups for patch restoration.  Furthermore, Levin \etal \cite{Levin2011Bounds} and  Chatterjee \etal \cite{Chatterjee2010IDD},  motivated external denoising~\cite{anwar2017category,anwar2017combined,luo2015adaptive,Yue2015CID} by showing 
that an image can be recovered with negligible error by selecting reference patches from a clean external database. However, all of the external algorithms are class-specific.
Recently, Schmidt \etal \cite{schmidt2014CSF} introduced a cascade of shrinkage fields (CSF) which integrated half-quadratic optimization and random-fields. Shrinkage aims to suppress smaller values (noise values) and learn mappings discriminatively. The CSF assumes the data fidelity term to be quadratic and that it has a discrete Fourier transform based closed-form solution.

Currently, due to the popularity of convolutional neural networks (CNNs), image denoising algorithms~\cite{zhang2017DnCNN,zhang2017IRCNN,lefkimmiatis2017NLNet, Burger2012MLP,schmidt2014CSF,anwar2017chaining} have achieved a performance boost. Notable denoising neural networks, DnCNN \cite{zhang2017DnCNN}, and IrCNN \cite{zhang2017IRCNN} predict the residue present in the image instead of the denoised image as the input to the loss function is ground truth noise as compared to the original clean image. Both networks achieved better results despite having a simple architecture where repeated blocks of convolutional, batch normalization and ReLU activations are used. Furthermore,  IrCNN \cite{zhang2017IRCNN} and DnCNN \cite{zhang2017DnCNN} are dependent on blindly predicted noise \ie without taking into account the underlying structures and textures of the noisy image.


Another essential image restoration framework is Trainable Nonlinear Reaction-Diffusion (TRND) \cite{chen2017TNRD} which uses a field-of-experts prior \cite{roth2009fields} into the deep neural network for a specific number of inference steps by extending the non-linear diffusion paradigm into a profoundly trainable parametrized linear filters and the influence functions.  Although the results of TRND are favorable, 
the model requires a significant amount of data to learn the parameters and influence functions as well as overall fine-tuning, hyper-parameter determination, and stage-wise training.  Similarly, non-local color net (NLNet) \cite{lefkimmiatis2017NLNet} was motivated by non-local self-similar (NSS) priors which employ non-local self-similarity coupled with discriminative learning. NLNet improved upon the traditional methods; but, it lags in performance compared to most of the CNNs \cite{zhang2017IRCNN,zhang2017DnCNN} due to the adaptaton of NSS priors, as it is unable to find the analogs for all the patches in the image.


Building upon the success of  DnCNN~\cite{zhang2017DnCNN}, Jiao \etal proposed a network consisting of two stacked subnets, named  ``FormattingNet''  and  ``DiffResNet''  respectively. The architecture of both networks is similar, and the difference lies in the loss layers used. The first subnet employs total variational and perceptual loss while the second one uses $\ell_2$ loss. The overall model is named as FormResNet and improves upon~\cite{zhang2017IRCNN,zhang2017DnCNN} by a small margin. Lately, Bae \etal~\cite{bae2017beyond} employed persistent homology analysis \cite{edelsbrunner2008persistent} via wavelet transformed domain to learn the features in CNN denoising. The performance of the model is marginally better compared to \cite{zhang2017DnCNN,jiao2017formresnet}, which can be attributed to a large number of feature maps employed rather than the model itself. Recently,  Anwar~\etal introduced CIMM, a deep denoising CNN architecture, composed of identity mapping modules~\cite{anwar2017chaining}. The network learns features in cascaded identity modules using dilated kernels and uses self-ensemble to boost performance. CIMM improved upon all the previous CNN models~\cite{zhang2017DnCNN,jiao2017formresnet}.




\begin{figure*}
\begin{center}
\includegraphics[clip, trim=1.5cm 3.5cm 1.5cm 3cm, width=\textwidth]{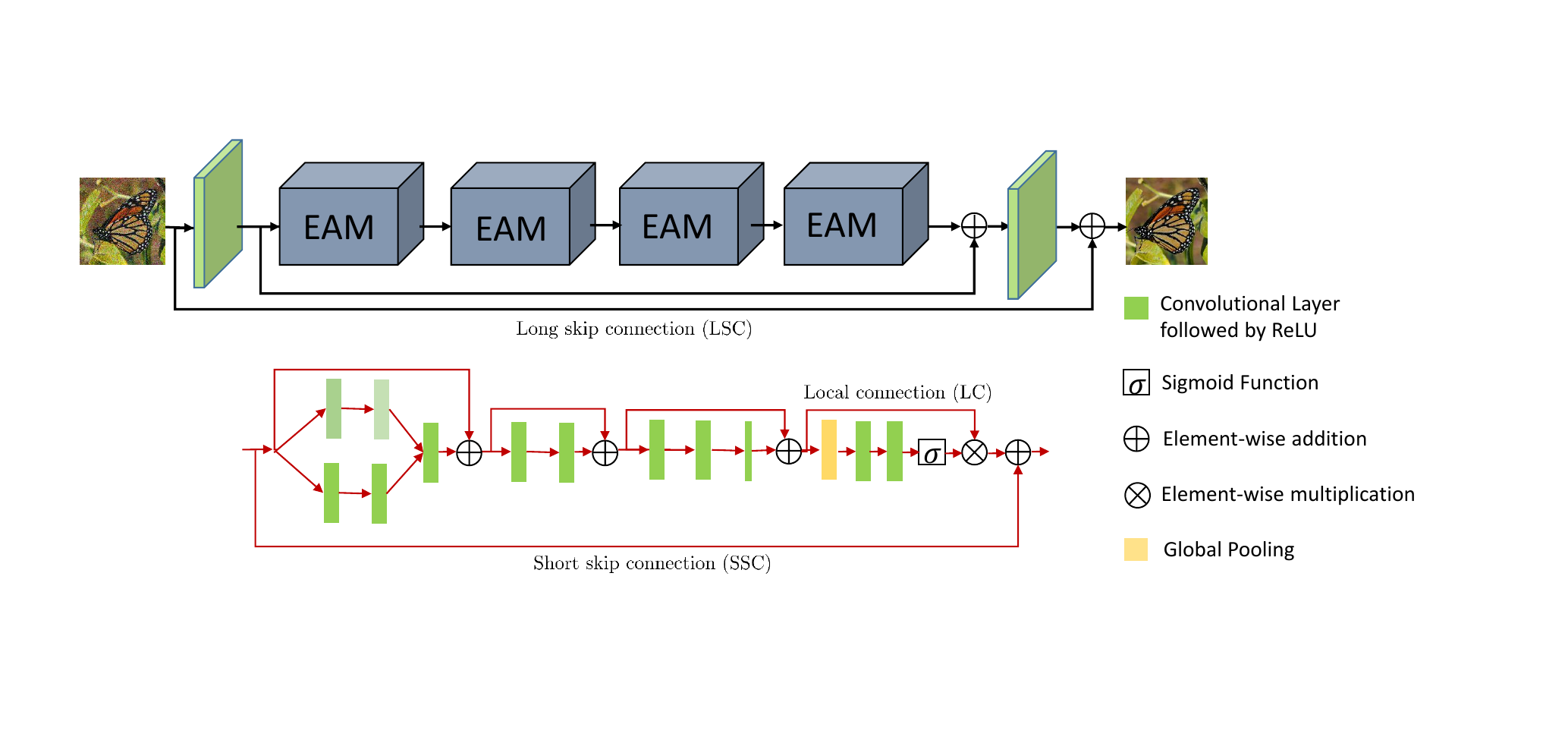}\\ 
\end{center}
\vspace*{-3mm}
\caption {The architecture of the proposed network. Different green colors of the conv layers denote different dilations while the smaller size of the conv layer means the kernel is $1 \times 1$.  The second row shows the architecture of each EAM.}
\label{fig:net_architecture}
\vspace*{-5mm}
\end{figure*}

Recently, many algorithms focused on blind denoising on real-noisy images~\cite{plotz2018N3Net,guo2018CBDnet,brooks2019UPI}. The algorithms~\cite{zhang2017IRCNN,zhang2017DnCNN,jiao2017formresnet} benefitted from the modeling capacity of CNNs and have shown the ability to learn a single-blind denoising model; however, the denoising performance is limited, and the results are not satisfactory on real photographs.  Generally speaking, real-noisy image denoising is a two-step process: the first involves noise estimation while the second addresses
non-blind denoising. Noise clinic (NC)~\cite{lebrun2015NC} estimates the noise model dependent on signal and frequency followed by denoising the image using non-local Bayes (NLB). In comparison, Zhang~\etal~\cite{zhang2018ffdnet} proposed a non-blind Gaussian denoising network, termed FFDNet that can produce satisfying results on some of the real noisy images; however, it requires manual intervention to select high noise-level.

Very recently, CBDNet~\cite{guo2018CBDnet} trains a blind denoising model for real photographs. CBDNet~\cite{guo2018CBDnet} is composed of two subnetworks: noise estimation and non-blind denoising.  CBDNet~\cite{guo2018CBDnet} also incorporated multiple losses, is engineered to train on real-synthetic noise and real-image noise and enforces a higher noise standard deviation for low noise images. Furthermore,~\cite{guo2018CBDnet,zhang2018ffdnet} may require manual intervention to improve results. On the other hand, we present an end-to-end architecture that learns the noise and produces results on real noisy images without requiring separate subnets or manual intervention.

\section{CNN Denoiser}
\subsection{Network Architecture}
Our model is composed of three main modules \ie feature extraction, feature learning residual on the residual module, and reconstruction, as shown in Figure \ref{fig:net_architecture}.  Let us consider $x$ is a noisy input image and $\hat{y}$ is the denoised output image.  Our feature extraction module is composed of only one convolutional layer to extract initial features $f_0$ from the noisy input:

\begin{equation}
f_0 = M_e(x),
\label{eq:extraction}    
\end{equation}
where $M_e(\cdot)$ performs convolution on the noisy input image. Next, $f_0$ is passed on to the feature learning residual on the residual module, termed as $M_{fl}$, 
\begin{equation}
f_r = M_{fl}(f_0),
\label{eq:fl}
\end{equation}  
where $f_r$ are the learned features and $M_{fl}(\cdot)$ is the main feature learning residual on the residual component, composed of enhancement attention modules (EAM) that are cascaded together as shown in Figure~\ref{fig:net_architecture}. Our network has small depth, but provides a wide receptive field through kernel dilation in each EAM initial two branch convolutions. The output features of the final layer are fed to the reconstruction module, which is again composed of one convolutional layer.

\begin{equation}
\hat{y} = M_r(f_r),
\label{eq:r}
\end{equation}  
where
$M_r(\cdot)$ denotes the reconstruction layer. 

There are several choices available as loss function for optimization such as  $\ell_2$~\cite{zhang2017DnCNN,zhang2017IRCNN,anwar2017chaining}, perceptual loss~\cite{jiao2017formresnet,guo2018CBDnet}, total variation loss~\cite{jiao2017formresnet} and asymmetric loss~\cite{guo2018CBDnet}. Some networks~\cite{jiao2017formresnet,guo2018CBDnet} employs more than one loss to optimize the model, contrary to  earlier networks, we only employ one loss \ie $\ell_1$. Now, given a batch of $N$ training pairs, $\{x_i, y_i\}_{i=1}^N$, where $x$ is the noisy input and $y$ is the ground truth, the aim is to minimize the $\ell_1$ loss function as 
\begin{equation}
L(\mathcal{W}) = \frac{1}{N} \sum_{i=1}^N||\text{RIDNet}(x_i) - y_i||_1,
\label{eq:l1_loss}
\end{equation}
where RIDNet($\cdot$) is our network and $\mathcal{W}$ denotes the set of all the network parameters learned. Our feature extraction $M_e$ and reconstruction module $M_r$ resemble the previous algorithms~\cite{dong2016SRCNN,anwar2017chaining}. We now focus on the feature learning residual on the residual block, and feature attention.

\subsection{Feature learning Residual on the Residual}
\label{ss:fl_rir}
In this section, we provide more details on the enhancement attention modules that uses a Residual on the Residual structure with  local skip and short skip connections.  Each EAM is further composed of $D$ blocks followed by feature attention. Due to the residual on the residual architecture, very deep networks are now possible that improve denoising performance; however, we restrict our model to four EAM modules only. The first part of EAM covers the full receptive field of input features, followed by learning on the features; then the features are compressed for speed, and finally a feature attention module enhances the weights of important features from the maps. The first part of EAM is realized using a novel merge-and-run unit as shown in Figure~\ref{fig:net_architecture} second row. The input features branched and are passed through two dilated convolutions, then concatenated and passed through another convolution. Next, the features are learned using a residual block of two convolutions while compression is achieved by an enhanced residual block (ERB) of three convolutional layers. The last layer of ERB flattens the features by applying a $1 \times 1$ kernel. Finally, the output of the feature attention unit is added to the input of EAM.

In image recognition, residual blocks \cite{he2016deep} are stacked together to construct a network of more than 1000 layers. Similarly, in image superresolution, EDSR \cite{lim2017EDSR} stacked the residual blocks and used long skip connections (LSC) to form a very deep network. However, to date, very deep networks have not been investigated for denoising. Motivated by the success of \cite{zhang2018RCAN}, we introduce the residual on the residual as a basic module for our network to construct deeper systems. Now consider the m-th module of the EAM is given as
\begin{equation}
f_m = EAM_m(EAM{m-1}(\cdots (M_0(f_0)) \cdots)),
\label{eq:rir_block}
\end{equation}
where $f_m$ is the output of the $EAM_m$ feature learning module, in other words $f_m = EAM_m(f_{m-1})$. The output of each EAM is added to the input of the group as $f_m = f_m+f_{m-1}$. We have observed that simply cascading the residual modules will not achieve better performance, instead we add the input of the feature extractor module to the final output of the stacked modules as 
\begin{equation}
f_g = f_0  + M_{fl}(\mathcal{W}_{w,b}),
\label{eq:residual_block}
\end{equation}
where $\mathcal{W}_{w,b}$ are the weights and biases learned in the group. This addition \ie LSC, eases the flow of information across groups. $f_g$ is passed to reconstruction layer to output the same number of channels as the input of the network. Furthermore, we use another long skip connection to add the input image to the network output \ie $\hat{y} = M_r(f_g)+x$, in order to learn the residual (noise) rather than the denoised image, as this technique helps in faster learning as compared to learning original image due to the sparse representation of the noise.

\subsubsection{Feature Attention}
\begin{figure}
\begin{center}
\includegraphics[clip, trim=5cm 8cm 5cm 6cm, width=0.5\textwidth]{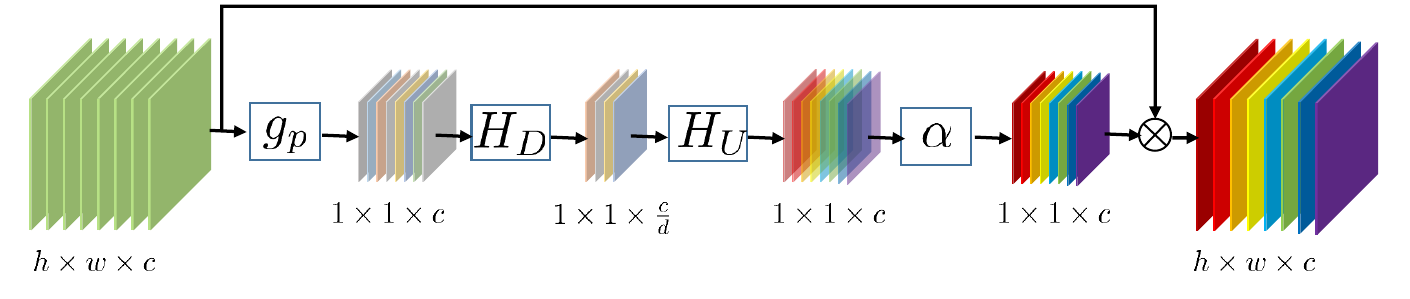}\\ 
\end{center}
\caption {The feature attention mechanism for selecting the essential features.}
\label{fig:net_eam}
\end{figure}

This section provides information about the feature attention mechanism. Attention \cite{xu2015show} has been around for some time; however, it has not been employed in image denoising. Channel features in image denoising methods are treated equally, which is not appropriate for many cases.  To exploit and learn the critical content of the image, we focus  attention on the relationship between the channel features; hence the name: feature attention (see Figure~\ref{fig:net_eam}).

An important question here is how to generate attention differently for each channel-wise feature. Images generally can be considered as having low-frequency regions (smooth or flat areas), and high-frequency regions (\eg, lines edges and texture). 
As convolutional layers exploit local information only and are unable to utilize global contextual information, we first employ global average pooling to express the statistics denoting the whole image, other options for aggregation of the features can also be explored to represent the image descriptor. Let $f_c$ be the output features of the last convolutional layer having $c$ feature maps of size $h \times w$; global average pooling will reduce the size from $h \times w \times c$ to $1 \times 1 \times c$ as:

\begin{equation}
g_p = \frac{1}{h \times w} \sum_{i=1}^h \sum_{i=1}^w f_c(i,j),
\label{eq:GAP}
\end{equation}
where $f_c(i,j)$ is the feature value at position $(i,j)$ in the feature maps. 

Furthermore as investigated in \cite{hu2018squeeze}, we propose a self-gating mechanism to capture the channel dependencies from the descriptor retrieved by  global average pooling. According to \cite{hu2018squeeze}, the mentioned mechanism must learn the nonlinear synergies between channels as well as mutually-exclusive relationships. Here, we employ soft-shrinkage and sigmoid functions to implement the gating mechanism. Let us consider $\delta$, and $\alpha$ are the soft-shrinkage and sigmoid operators, respectively. Then the gating mechanism is 
\begin{equation}
r_c =  \alpha(H_U(\delta(H_D(g_p)))),
\label{eq:gating}
\end{equation}
where $H_D$ and $H_U$ are the channel reduction and channel upsampling operators, respectively. The output of the global pooling layer $g_p$ is convolved with a downsampling Conv layer, activated by the soft-shrinkage function. To differentiate the channel features, the output is then fed into an upsampling Conv layer followed by sigmoid activation.   Moreover, to compute the statistics, the output of the sigmoid ($r_c$) is adaptively rescaled by the input $f_c$ of the channel features as 
\begin{equation}
\hat{f}_c = r_c\times f_c
\label{eq:rescale_CA}
\end{equation}

\subsection{Implementation}

\begin{table*}
\centering
\begin{tabular}{l||c|c|c|c|c|c|c|c|c}
\hline\hline
 Long skip connection (LSC)     &      & \ch   & 	  &  \ch &       &        &       &  \ch   &\ch \\ 
  Short skip connection (SSC)   &      &  	   & \ch  &  \ch &       &        & \ch   &  \ch   &\ch \\  
  Long connection (LC)          &      &  	   & 	  &      &       & \ch          & \ch   &        &\ch \\
 Feature attention (FA)         &      &  	   & 	  &      &  \ch  & \ch       & \ch   & \ch    &\ch \\  \hline
 PSNR (in dB)  &28.45 & 28.77 &28.81 &28.86 & 28.52 &28.85  & 28.86 & 28.90 &28.96 \\  \hline \hline
\end{tabular}
\caption{ Investigation of skip connections and feature attention. The best result in PSNR (dB) on values on BSD68~\cite{roth2009fields} in 2$\times10^{5}$ iterations is presented.}
\label{table:ablation}
\vspace*{-5mm}
\end{table*}
Our proposed model contains four EAM blocks. The kernel size for each convolutional layer is set to $3 \times 3$, except the last Conv layer in the enhanced residual block and those of the features attention units, where the kernel size is $1 \times 1$. Zero padding is used for $3 \times 3$ to achieve the same size outputs feature maps. The number of channels for each convolutional layer is fixed at 64, except for feature attention downscaling. A factor of 16 reduces these Conv layers; hence having only four feature maps. The final convolutional layer either outputs three or one feature maps depending on the input. As for running time, our method takes about 0.2 second to process a $512 \times 512$ image.

\section{Experiments}

\subsection{Training settings}
To generate noisy synthetic images, we employ BSD500~\cite{Martin2001BSD}, DIV2K~\cite{agustsson2017ntire}, and MIT-Adobe FiveK~\cite{bychkovsky2011learning}, resulting in 4k images while for real noisy images, we use cropped patches of $512 \times 512$ from SSID~\cite{abdelhamed2018high}, Poly~\cite{xu2018real}, and RENOIR~\cite{anaya2018renoir}. Data augmentation is performed on training images, which includes random rotations of 90$^{\circ}$, 180$^{\circ}$, 270$^{\circ}$ and flipping horizontally. In each training batch, 32 patches are extracted as inputs with a size of $80 \times 80$.  Adam~\cite{kingma2014adam} is used as the optimizer with default parameters. The learning rate is initially set to $10^{-4}$ and then halved after $10^{5}$ iterations. The network is implemented in the Pytorch~\cite{paszke2017automatic} framework and trained with an Nvidia Tesla V100 GPU. Furthermore, we use PSNR as evaluation metric.

\subsection{Ablation Studies}

\subsubsection{Influence of the skip connections}
Skip connections play a crucial role in our network. Here, we demonstrate the effectiveness of the skip connections. Our model is composed of three basic types of connections which includes long skip connection (LSC),  short skip connections (SSC), and local connections (LC). Table~\ref{table:ablation} shows the average PSNR for the BSD68~\cite{roth2009fields} dataset. The highest performance is obtained when all the skip connections are available while the performance is lower when any connection is absent. We also observed that increasing the depth of the network in the absence of skip connections does not benefit performance.

\subsubsection{Feature-attention}
Another important aspect of our network is feature attention. Table~\ref{table:ablation} compares the PSNR values of the networks with and without feature attention. The results support our claim about the benefit of using feature attention. Since the inception of DnCNN~\cite{zhang2017DnCNN}, the CNN models have matured, and further performance improvement requires the careful design of blocks and rescaling of the feature maps. The two mentioned characteristics are present in our model in the form of feature-attention and the skip connections.


\subsection{Comparisons}
We evaluate our algorithm using the Peak Signal-to-Noise Ratio (PSNR) index as the error metric and compare against many state-of-the-art competitive algorithms which include traditional methods \ie CBM3D~\cite{dabov2007CBM3D}, WNNM~\cite{Gu2014WNN},  EPLL~\cite{Zoran2011EPLL}, CSF~\cite{schmidt2014CSF} and CNN-based denoisers \ie MLP~\cite{Burger2012MLP},  TNRD~\cite{chen2017TNRD}, DnCNN~\cite{zhang2017DnCNN}, IrCNN~\cite{zhang2017IRCNN}, CNLNet~\cite{lefkimmiatis2017NLNet}, FFDNet~\cite{zhang2018ffdnet} and CBDNet~\cite{guo2018CBDnet}. To be fair in comparison, we use the default setting of the traditional methods provided by the corresponding authors.

\subsubsection{Test Datasets}
In the experiments, we test four noisy real-world datasets \ie RNI15~\cite{lebrun2015NC}, DND~\cite{plotz2017benchmarking}, Nam~\cite{nam2016holistic} and SSID~\cite{abdelhamed2018high}. Furthermore, we prepare three synthetic noisy datasets from the widely used 12 classical images, BSD68~\cite{roth2009fields} color and gray 68 images for testing. We corrupt the clean images by additive white Gaussian noise using noise sigma of 15, 25 and 50 standard deviations.
\begin{itemize}


\begin{table*}
\centering
\begin{tabular}{c|c|c|c|c|c|c|c|c|c|c}
\hline\hline
 Noise & \multicolumn{9}{c}{Methods} \\ 
 Level& BM3D    & WNNM  	& EPLL	  & TNRD   & DenoiseNet & DnCNN   & IrCNN   & NLNet     &FFDNet	& Ours	  \\\hline 
 15    & 31.08   & 31.32 	& 31.19	  & 31.42  & 31.44 	    & 31.73   &31.63    & 31.52	    &31.63	&\textbf{31.81} \\ 
 25    & 28.57   & 28.83	& 28.68	  & 28.92  & 29.04 	    & 29.23   &29.15    & 29.03 	&29.23	&\textbf{29.34} \\  
 50    & 25.62   & 25.83	& 25.67	  & 26.01  & 26.06  	& 26.23   &26.19    & 26.07	    &26.29	&\textbf{26.40} \\ \hline \hline
\end{tabular}
\caption{The similarity between the denoised and the clean images of BSD68 dataset~\cite{roth2009fields} for our method and competing measured in terms of average PSNR for $\sigma$=15, 25, and 50 on grayscale images.}
\label{table:BSD68_grayscale}
\end{table*}

\item RNI15~\cite{lebrun2015NC} provides 15 real-world noisy images. Unfortunately, the clean images are not given for this dataset; therefore, only the qualitative comparison is presented for this dataset.

\item Nam~\cite{nam2016holistic} comprises of 11 static scenes and the corresponding noise-free images obtained by the mean of 500 noisy images of the same scene. The size of the images are enormous; hence, we cropped the images in $512 \times 512$ patches and randomly selected 110 from those for testing.

\item DnD is recently proposed by Plotz~\etal~\cite{plotz2017benchmarking}  which originally contains 50 pairs of real-world noisy and noise-free scenes. The scenes are further cropped into patches of size  $512 \times 512$ by the providers of the dataset which resulted in 1000 smaller images. The near noise-free images are not publicly available, and the results (PSNR/SSIM) can only be obtained through the online system introduced by~\cite{plotz2017benchmarking}.

\item SSID~\cite{abdelhamed2018high} (Smartphone Image Denoising Dataset) is recently introduced.  The authors have collected 30k real noisy images and their corresponding clean images; however, only 320 images are released for training and 1280 images pairs for validation, as testing images are not released yet. We will use the validation images for testing our algorithm and the competitive methods. 
\end{itemize}

\subsubsection{Grayscale noisy images}
\begin{table}
\centering
\begin{tabular}{|l||ccc|}\hline 
Methods    &   $\sigma$ = 15 & $\sigma$ = 25    &   $\sigma$ = 50 \\ \hline \hline
BM3D~\cite{Dabov2007BM3D}    & 32.37 &   29.97  &   26.72 \\
WNNM~\cite{Gu2014WNN}        & 32.70 &   30.26  &   27.05 \\
EPLL~\cite{Zoran2011EPLL}    & 32.14 &   29.69  &   26.47 \\
MLP~\cite{Burger2012MLP}     & -     &   30.03  &   26.78 \\
CSF~\cite{schmidt2014CSF}    & 32.32 &   29.84  &    -    \\
TNRD~\cite{chen2017TNRD}     & 32.50 &   30.06  &   26.81 \\
DnCNN~\cite{zhang2017DnCNN}  & 32.86 &   30.44  &   27.18 \\
IrCNN ~\cite{zhang2017IRCNN} & 32.77 &   30.38  &   27.14 \\ 
FFDNet~\cite{zhang2018ffdnet}& 32.75 &   30.43  &   27.32 \\
Ours                         & \textbf{32.91} &   \textbf{30.60}  &   \textbf{27.43} \\ \hline \hline
\end{tabular}
\caption{The quantitative comparison between denoising algorithms on 12 classical images, (in terms of PSNR). The best results are highlighted as bold.}
\label{table:classical}
\vspace*{-3mm}
\end{table}

In this subsection, we evaluate our model on the noisy grayscale images corrupted by spatially invariant additive white Gaussian noise. We compare against nonlocal self-similarity representative models \ie BM3D~\cite{Dabov2007BM3D} and WNNM~\cite{Gu2014WNN}, learning based methods \ie EPLL, TNRD~\cite{chen2017TNRD}, MLP~\cite{Burger2012MLP}, DnCNN~\cite{zhang2017DnCNN}, IrCNN~\cite{zhang2017IRCNN}, and CSF~\cite{schmidt2014CSF}.  In Tables~\ref{table:classical} and~\ref{table:BSD68_grayscale}, we present the PSNR values on Set12 and BSD68. It is to be remembered here that BSD500~\cite{Martin2001BSD} and BSD68~\cite{roth2009fields} are two disjoint sets.  Our method outperforms all the competitive algorithms on both datasets for all noise levels; this may be due to the larger receptive field as well as better modeling capacity.

\subsubsection{Color noisy images}
\begin{figure}
\begin{center}
\begin{tabular}{c@{ }  c@{ } c}
\includegraphics[trim={1.5cm 5.15cm  1.5cm  5cm },clip,width=.145\textwidth,valign=t]{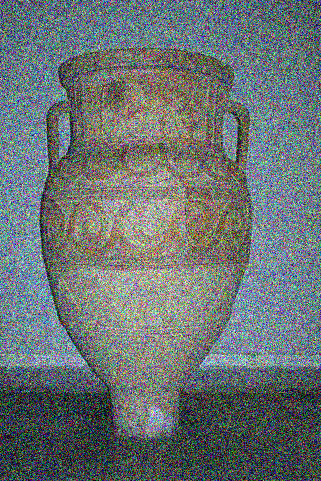}&   
\includegraphics[trim={1.5cm 5.15cm  1.5cm  5cm },clip,width=.145\textwidth,valign=t]{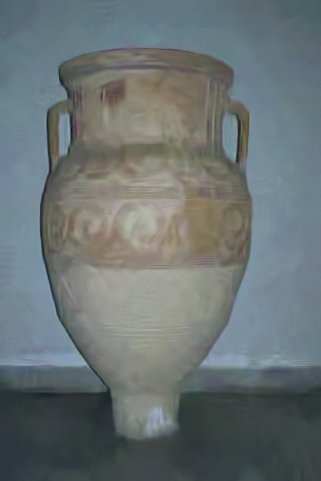}& 
\includegraphics[trim={1.5cm 5.15cm  1.5cm  5cm},clip,width=.145\textwidth,valign=t]{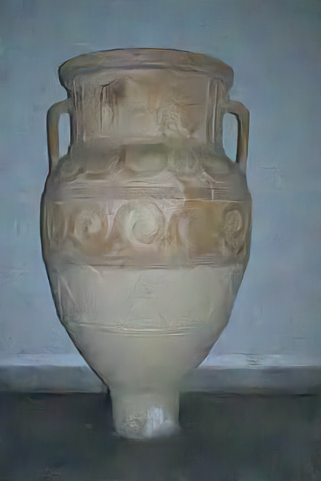}\\
& 31.68dB& 32.21dB\\
   Noisy & BM3D~\cite{dabov2007CBM3D}    & IRCNN~\cite{zhang2017IRCNN} \\

\includegraphics[trim={1.5cm 5.15cm  1.5cm  5cm },clip,width=.145\textwidth,valign=t]{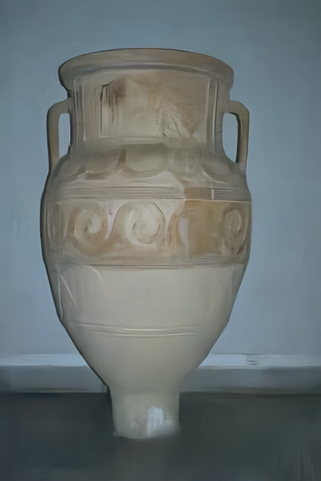}&
 \includegraphics[trim={1.5cm 5.15cm  1.5cm  5cm },clip,width=.145\textwidth,valign=t]{images/BSD68/227092_Ours}&
 \includegraphics[trim={1.5cm 5.15cm  1.5cm  5cm },clip,width=.145\textwidth,valign=t]{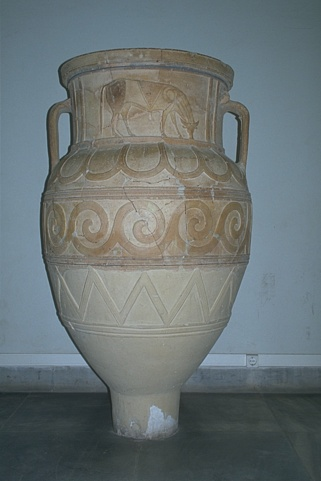}\\
 32.33dB& 32.84dB & \\
   DnCNN~\cite{zhang2017DnCNN} & Ours    & GT \\

\end{tabular}
\end{center}
\vspace*{-3.5mm}
\caption{Denoising performance of our RIDNet versus state-of-the-art  methods on a color images from~\cite{roth2009fields} for $\sigma_n = 50$}
\label{fig:BSD68_color}
\vspace*{-5mm}
\end{figure}

\begin{table*}
\centering
\begin{tabular}{c|c|c|c|c|c|c|c|c}
\hline \hline
Noise  & \multicolumn{7}{c}{Methods} \\ 
Levels & CBM3D~\cite{dabov2007CBM3D}       & MLP~\cite{Burger2012MLP}    & TNRD~\cite{chen2017TNRD}  & DnCNN~\cite{zhang2017DnCNN}  & IrCNN~\cite{zhang2017IRCNN}  & CNLNet~\cite{lefkimmiatis2017NLNet} &FFDNet~\cite{zhang2018ffdnet}& Ours   \\ \hline
 15    &  33.50      & -      & 31.37 & 33.89   					& 33.86 & 33.69	&33.87 & \textbf{34.01}\\  
 25    &  30.69      & 28.92  & 28.88 & 31.33   	& 31.16 & 30.96 &31.21 & \textbf{31.37} \\  
 50    &  27.37      & 26.00  & 25.94 & 27.97   					& 27.86 & 27.64	&27.96 & \textbf{28.14}\\ \hline  \hline                
\end{tabular}
\caption{Performance comparison between our network and existing state-of-the-art algorithms on the color version of the BSD68 dataset~\cite{roth2009fields}.}
\label{table:BSD68_color}
\end{table*}
Next, for noisy color image denoising, we keep all the parameters of the network similar to the grayscale model, except the first and last layer are changed to input and output three channels rather than one. Figure~\ref{fig:BSD68_color} presents the visual comparison and Table~\ref{table:BSD68_color} reports the PSNR numbers between our methods and the alternative algorithms. Our algorithm consistently outperforms all the other techniques published in Table~\ref{table:BSD68_color} for CBSD68 dataset~\cite{roth2009fields}.  Similarly, our network produces the best perceptual quality images as shown in Figure~\ref{fig:BSD68_color}. A closer inspection on the vase reveals that our network generates textures closest to the ground-truth with fewer artifacts and more details.

\subsubsection{Real-World noisy images}
To further assess the practicality of our model, we employ a real noise dataset.  The evaluation is difficult because of the unknown level of noise, the various noise sources such as shot noise, quantization noise \etc, imaging pipeline \ie image resizing, lossy compression \etc Furthermore, the noise is spatially variant (non-Gaussian) and also signal dependent; hence, the assumption that noise is spatially invariant, employed by many algorithms does not hold for real image noise. Therefore, real-noisy images evaluation determines the success of the algorithms in real-world applications.

\begin{figure*}
\begin{center}
\begin{tabular}[b]{c@{ } c@{ }  c@{ } c@{ } c@{ } c@{ }	}
    \multirow{4}{*}{\includegraphics[width=.314\textwidth,valign=t]{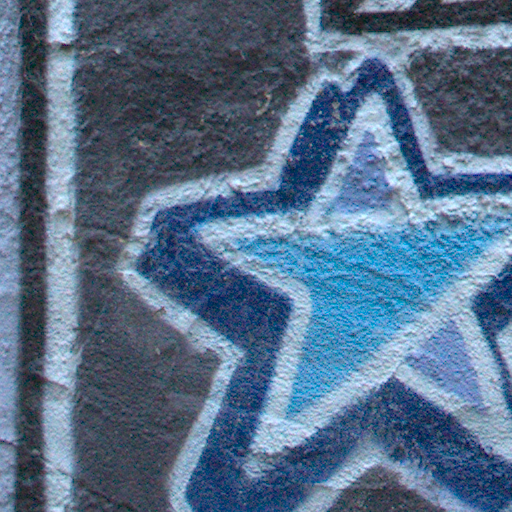}} &  
    \includegraphics[trim={3cm 3cm  3cm  3cm },clip,width=.133\textwidth,valign=t]{images/DnD/3_nosiy.png}&
  	\includegraphics[trim={3cm 3cm  3cm  3cm },clip,width=.13\textwidth,valign=t]{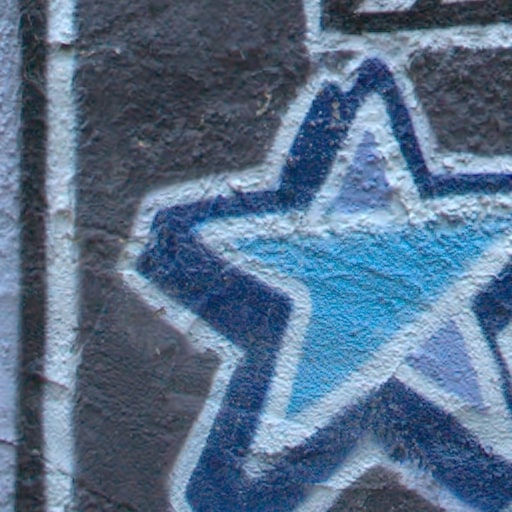}&   
    \includegraphics[trim={3cm 3cm  3cm  3cm },clip,width=.133\textwidth,valign=t]{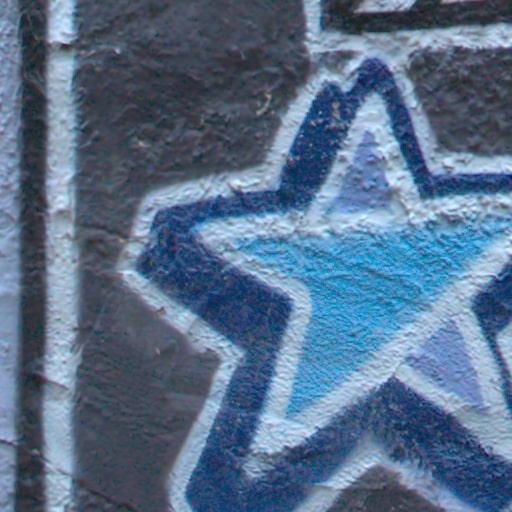}&
      	\includegraphics[trim={3cm 3cm  3cm  3cm },clip,width=.133\textwidth,valign=t]{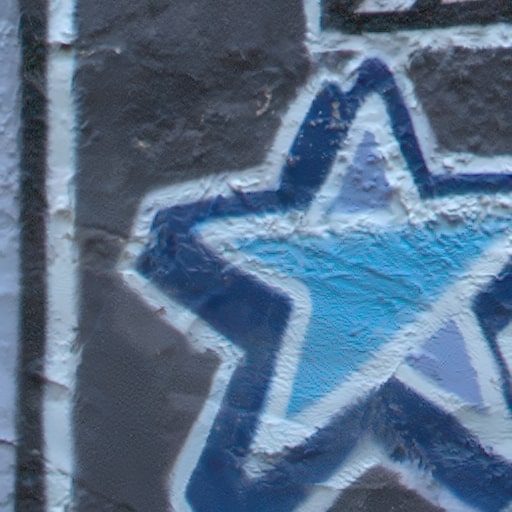}&
	\includegraphics[trim={3cm 3cm  3cm  3cm },clip,width=.133\textwidth,valign=t]{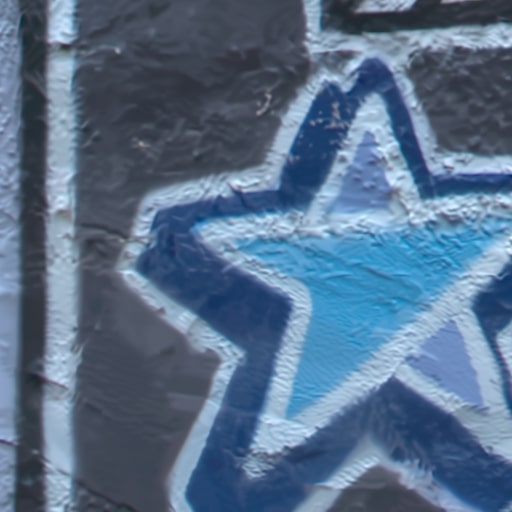}   
\\
    &       &30.896dB  & 29.98dB & 30.73dB & 29.42dB   \\
    & Noisy &CBM3D~\cite{Dabov2007BM3D}  & WNNM~\cite{Gu2014WNN}    & NC~\cite{lebrun2015NC}      & TWSC~\cite{xu2018TWSC} \\

    &
    \includegraphics[trim={3cm 3cm  3cm  3cm },clip,width=.133\textwidth,valign=t]{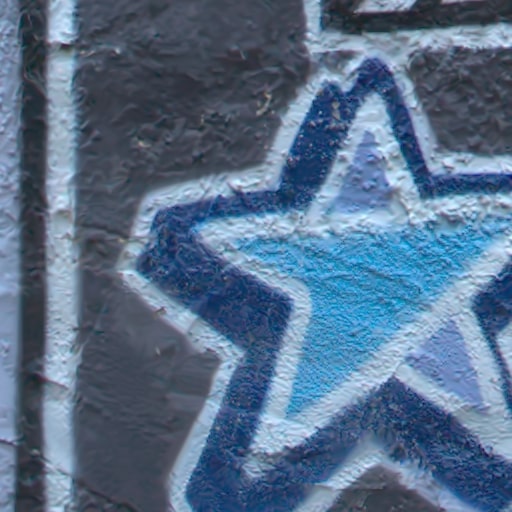}&
    \includegraphics[trim={3cm 3cm  3cm  3cm },clip,width=.133\textwidth,valign=t]{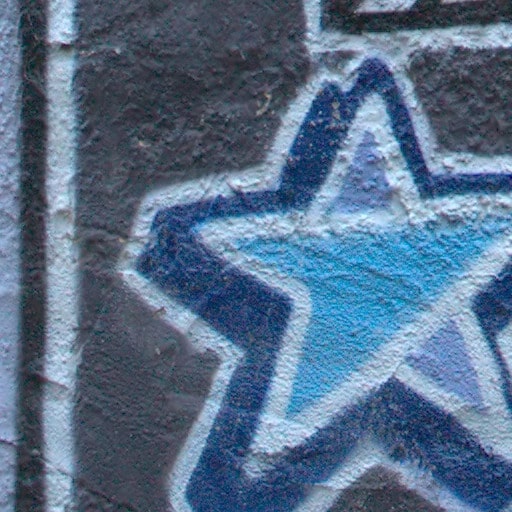}&
    \includegraphics[trim={3cm 3cm  3cm  3cm },clip,width=.133\textwidth,valign=t]{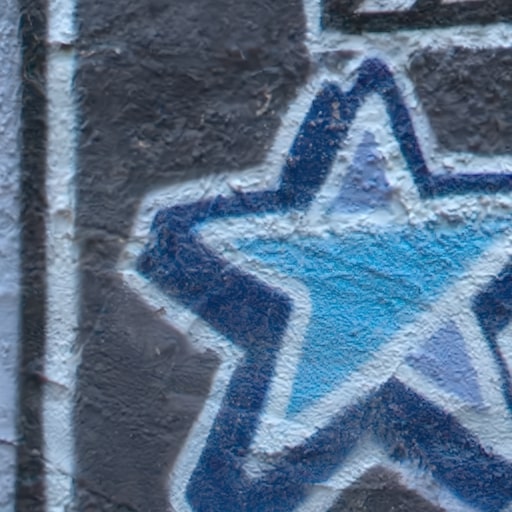}&  
    \includegraphics[trim={3cm 3cm  3cm  3cm },clip,width=.133\textwidth,valign=t]{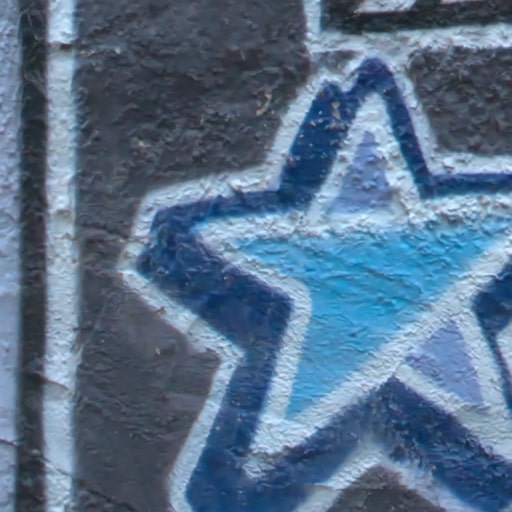}&   

     \includegraphics[trim={3cm 3cm  3cm  3cm },clip,width=.133\textwidth,valign=t]{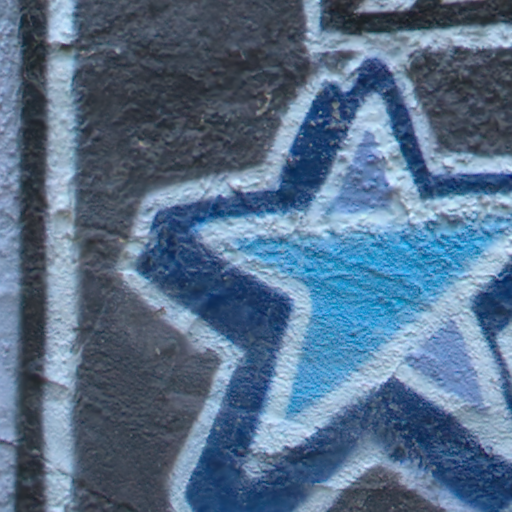}\\

     & 30.88dB & 28.43dB & 31.37dB & 31.06dB & \textbf{32.31dB}  \\
           Noisy Image  & MCWNNM~\cite{xu2017MCW}    & NI~\cite{NeatI}     &FFDNet~\cite{zhang2018ffdnet}       & CBDNet~\cite{guo2018CBDnet}     & RIDNet (Ours) \\
    
\end{tabular}
\end{center}
\vspace*{-4mm}
\caption{A real noisy example from DND dataset~\cite{plotz2017benchmarking} for comparison of our method against the state-of-the-art algorithms.}
\label{fig:DnD}
\vspace*{-5mm}
\end{figure*}

\paragraph{DnD:} 

\begin{table}
\centering
\begin{tabular}{l||ccc}
\hline \hline
Method      &Blind/Non-blind &PSNR &SSIM \\ \hline \hline
CDnCNNB~\cite{zhang2017DnCNN}  	&Blind 		&32.43 &0.7900\\
EPLL~\cite{Zoran2011EPLL} 		&Non-blind 	&33.51 &0.8244\\
TNRD~\cite{chen2017TNRD}  		&Non-blind 	&33.65 &0.8306\\
NCSR~\cite{dong2012NCSR}  		&Non-blind 	&34.05 &0.8351\\
MLP~\cite{Burger2012MLP}  		&Non-blind 	&34.23 &0.8331\\
FFDNet~\cite{zhang2018ffdnet}  	&Non-blind 	&34.40 &0.8474\\
BM3D~\cite{Dabov2007BM3D}  		&Non-blind 	&34.51 &0.8507\\
FoE~\cite{roth2009fields}  		&Non-blind 	&34.62 &0.8845\\
WNNM~\cite{Gu2014WNN}  		&Non-blind 	&34.67 &0.8646\\
NC~\cite{lebrun2015NC}	        &Blind	&35.43	&0.8841	\\
NI~\cite{NeatI}	    	&Blind	&35.11  &0.8778 \\
CIMM~\cite{anwar2017chaining}  		&Non-blind 	&36.04 &0.9136\\
KSVD~\cite{aharon2006ksvd}  		&Non-blind 	&36.49 &0.8978\\
MCWNNM~\cite{xu2017MCW}  		&Non-blind 	&37.38 &0.9294\\
TWSC~\cite{xu2018TWSC}        &Non-blind	&37.96 &0.9416\\
FFDNet+~\cite{zhang2018ffdnet}		&Non-blind  &37.61 &0.9415\\
CBDNet~\cite{guo2018CBDnet}		&Blind 		&38.06 &0.9421\\
RIDNET (Ours)		&Blind 		&\textbf{39.23} & \textbf{0.9526}\\ \hline \hline
                
\end{tabular}
\caption{The Mean PSNR and SSIM denoising results of state-of-the-art algorithms evaluated on the DnD sRGB images~\cite{plotz2017benchmarking}}
\label{table:DnD}
\vspace*{-5mm}
\end{table}

Table~\ref{table:DnD} presents the quantitative results (PSNR/SSIM) on the sRGB data for competitive algorithms and our method obtained from the online DnD benchmark website available publicly. The blind Gaussian denoiser DnCNN~\cite{zhang2017DnCNN} performs inefficiently and is unable to achieve better results than BM3D~\cite{Dabov2007BM3D} and WNNM~\cite{Gu2014WNN} due to the poor generalization of the noise during training.  Similarly, the non-blind Gaussian traditional denoisers are able to report limited performance, although the noise standard-deviation is provided. This may be due to the fact that these denoisers~\cite{Dabov2007BM3D,Gu2014WNN,Zoran2011EPLL} are tailored for AWGN only and real-noise is different in characteristics to synthetic noise. Incorporating feature attention and capturing the appropriate characteristics of the noise through a novel module 
means our algorithm leads by large margin \ie 1.17dB PSNR compared to the second performing method, CBDNet~\cite{guo2018CBDnet}. Furthermore, our algorithm only employs real-noisy images for training using only $\ell_1$ loss while CBDNet~\cite{guo2018CBDnet} uses many techniques such as multiple losses (\ie total variation, $\ell_2$ and asymmetric learning) and both real-noise as well as synthetically generated real-noise. As reported by the author of CBDNet~\cite{guo2018CBDnet}, it is able to achieve 37.72 dB with real-noise images only. Noise Clinic (NC)~\cite{lebrun2015NC} and Neat Image (NI)~\cite{NeatI} are the other two state-of-the-art blind denoisers other than~\cite{guo2018CBDnet}. NI~\cite{NeatI} is commercially available as a part of Photoshop and Corel PaintShop. Our network is able to achieve 3.82dB and 4.14dB more PSNR from NC~\cite{lebrun2015NC} and NI~\cite{NeatI}, respectively. 

\begin{figure}
\begin{center}
\begin{tabular}[b]{c@{}c@{}c@{}c} 
      
\includegraphics[width=.12\textwidth]{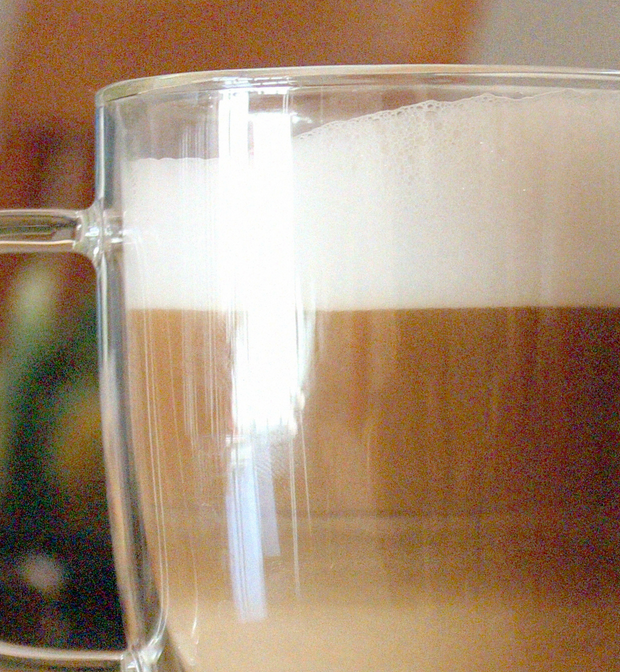}&
\includegraphics[width=.12\textwidth]{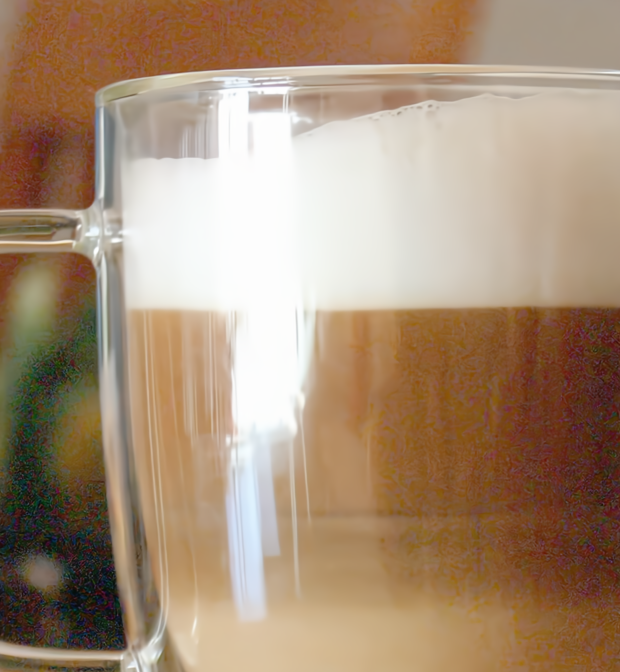}&   
\includegraphics[width=.12\textwidth]{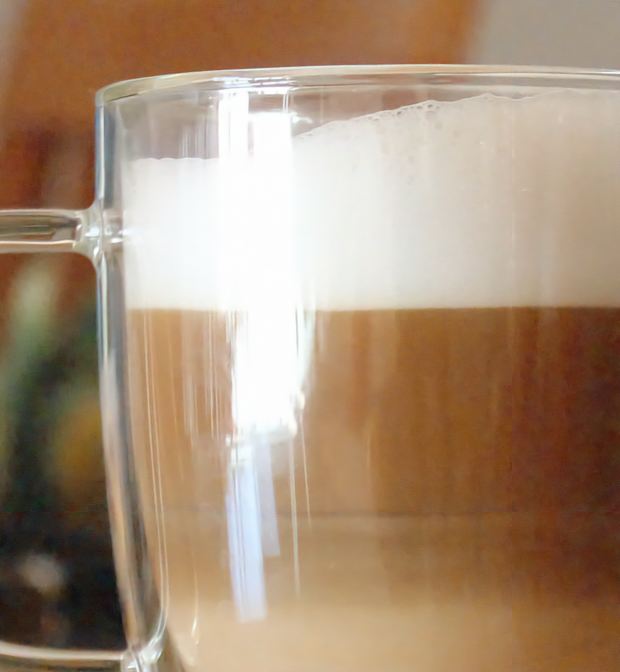}&
\includegraphics[width=.12\textwidth]{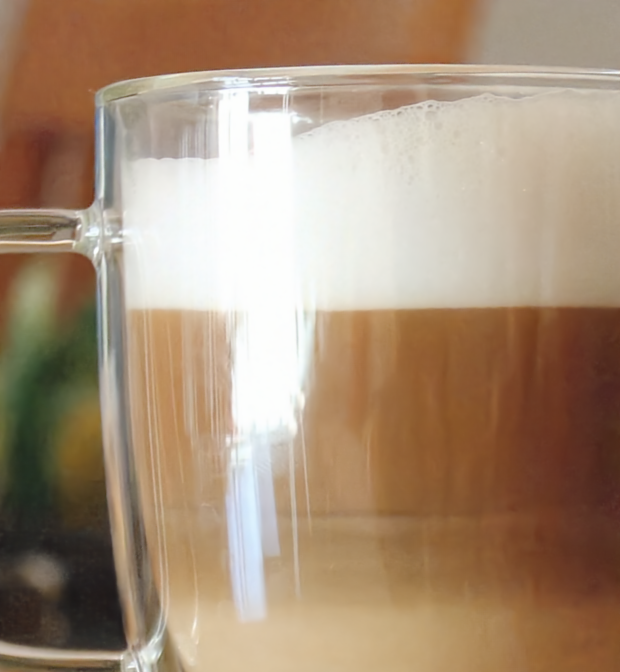}\\
Noisy & FFDNet & CBDNet & RIDNet\\
\end{tabular}
\end{center}
\vspace*{-2mm}
\caption{Comparison of our method against the other methods on a real image from RNI15~\cite{lebrun2015NC} benchmark containing spatially variant noise. }
\label{fig:NC}
\vspace*{-3mm}
\end{figure}

Next, we visually compare the result of our method with the competing methods on the denoised images provided by the online system of Plotz~\etal~\cite{plotz2017benchmarking} in Figure~\ref{fig:DnD}. The PSNR and SSIM values are also taken from the website. From Figure~\ref{fig:DnD}, it is clear that the methods of~\cite{guo2018CBDnet,zhang2018ffdnet,zhang2017DnCNN} perform poorly in removing the noise from the star and in some cases the image is over-smoothed, on the other hand, our algorithm can eliminate the noise while preserving the finer details and structures in the star image.

\begin{figure}
\begin{center}
\begin{tabular}[b]{c@{ } c@{ }  c@{ } c@{ } c@{ } c@{ }	}
    \multirow{2}{*}{\includegraphics[width=.15\textwidth,valign=t]{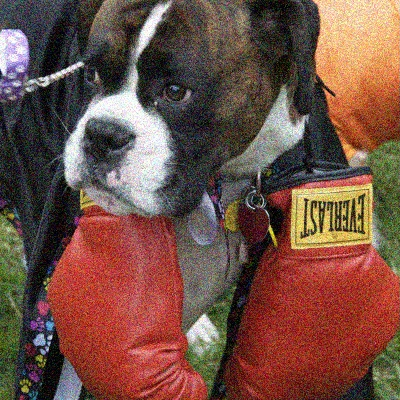}} &  
    \includegraphics[trim={5cm 6cm  1cm  1cm },clip,width=.10\textwidth,valign=t]{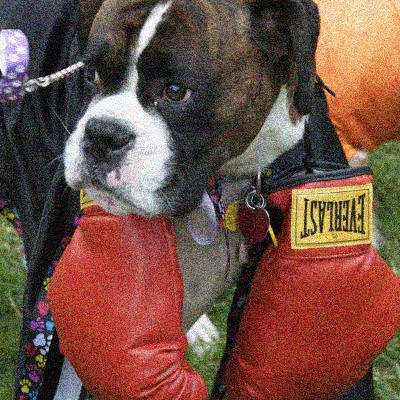}&
  	\includegraphics[trim={5cm 6cm  1cm  1cm },clip,width=.10\textwidth,valign=t]{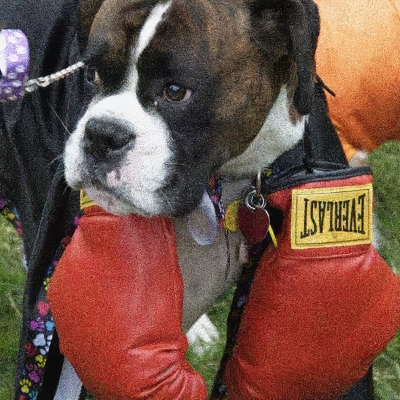}&   
    \includegraphics[trim={5cm 6cm  1cm  1cm },clip,width=.10\textwidth,valign=t]{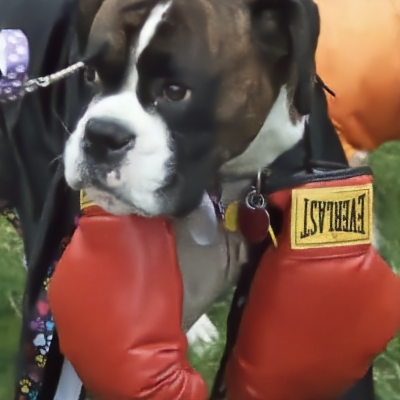}\\
    &
    \includegraphics[trim={1cm 2.2cm  7cm  8cm },clip,width=.10\textwidth,valign=t]{images/RNI15/RNI15_Dog_DnCNN.png}&
  	\includegraphics[trim={1cm 2.2cm  7cm  8cm },clip,width=.10\textwidth,valign=t]{images/RNI15/RNI15_Dog_FFD}&   
    \includegraphics[trim={1cm 2.2cm  7cm  8cm },clip,width=.102\textwidth,valign=t]{images/RNI15/RNI15_Dog_Our}\\
    Noisy    & DnCNN     & FFDNet       &  \textbf{Ours} \\
    
\end{tabular}
\end{center}
\vspace*{-2mm}
\caption{A real high noise example from RNI15 dataset~\cite{lebrun2015NC}. Our method is able to remove the noise in textured and smooth areas without introducing artifacts.}
\label{fig:NC_high}
\vspace*{-3mm}
\end{figure}

\paragraph{RNI15:} On RNI15~\cite{lebrun2015NC}, we provide qualitative images only as the ground-truth images are not available. Figure~\ref{fig:NC} presents the denoising results on a low noise intensity image. FFDNet~\cite{zhang2018ffdnet} and CBDNet~\cite{guo2018CBDnet} are unable to remove the noise in its totality as can been seen near the bottom left of handle and body of the cup image. On the contrary, our method is able to remove the noise without the introduction of any artifacts. We present another example from the RNI15 dataset~\cite{lebrun2015NC} with high noise in Figure~\ref{fig:NC_high}. CDnCNN~\cite{zhang2017DnCNN} and FFDNet~\cite{zhang2018ffdnet} produce results of limited nature as some noisy elements can be seen in the near the eye and gloves of the Dog image. In comparison, our algorithm recovers the actual texture and structures without compromising on the removal of noise from the images.

\begin{table}
\centering
\resizebox{\columnwidth}{!}{%
\begin{tabular}{l|c|c|c|c|c}
\hline\hline
  \multicolumn{6}{c}{Methods} \\ 
Datasets    &BM3D     &  DnCNN   & FFDNet	 & CBDNet    & Ours	  \\\hline 
Nam~\cite{nam2016holistic}         &37.30    &  35.55   & 38.7    & 39.01     &\textbf{39.09} \\ 
SSID~\cite{abdelhamed2018high}        &30.88    &  26.21   & 29.20   & 30.78     &\textbf{38.71} \\ \hline \hline
\end{tabular}}
\caption{The quantitative results (in PSNR (dB)) for the SSID~\cite{abdelhamed2018high} and Nam~\cite{nam2016holistic} datasets.}
\label{table:Nam_SSID}
\vspace*{-5mm}
\end{table}


\paragraph{Nam:} We present the average PSNR scores of the resultant denoised images in Table~\ref{table:Nam_SSID}. Unlike CBDNet~\cite{guo2018CBDnet}, which is trained on Nam~\cite{nam2016holistic} to specifically deal with the JPEG compression, we use the same network to denoise the Nam images~\cite{nam2016holistic} and achieve favorable PSNR numbers. Our performance in terms of PSNR is higher than any of the current state-of-the-art algorithms. Furthermore, our claim is supported by the visual quality of the images produced by our model as shown in Figure~\ref{fig:Nam}. The amount of noise present after denoising by our method is negligible as compared to CDnCNN and other counterparts. 

\begin{figure}
\begin{center}
\begin{tabular}[b]{c@{ }c@{ }c} 
      
\includegraphics[width=.15\textwidth]{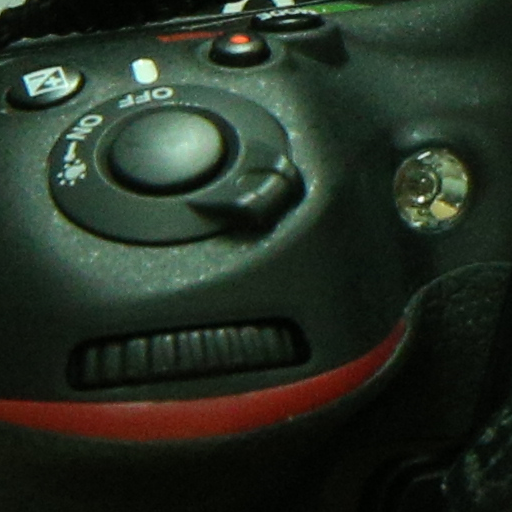}&   
\includegraphics[width=.15\textwidth]{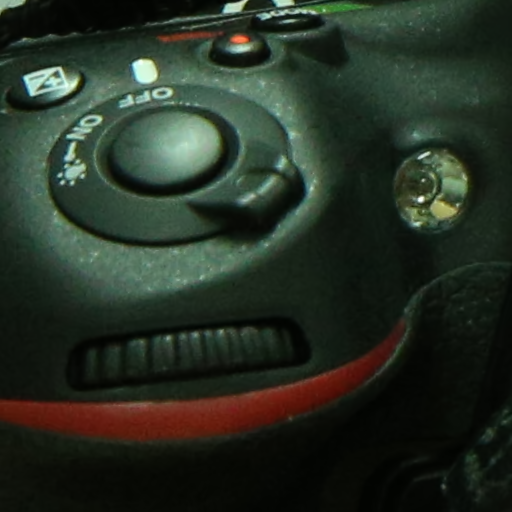}&
\includegraphics[width=.15\textwidth]{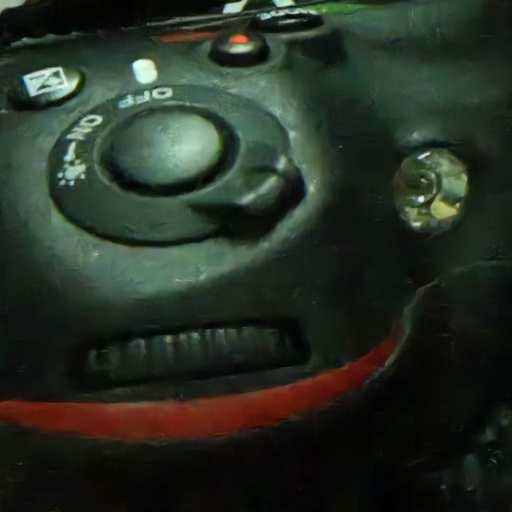}\\
Noisy & CBM3D (39.13) & IRCNN (33.73)\\
\includegraphics[width=.15\textwidth]{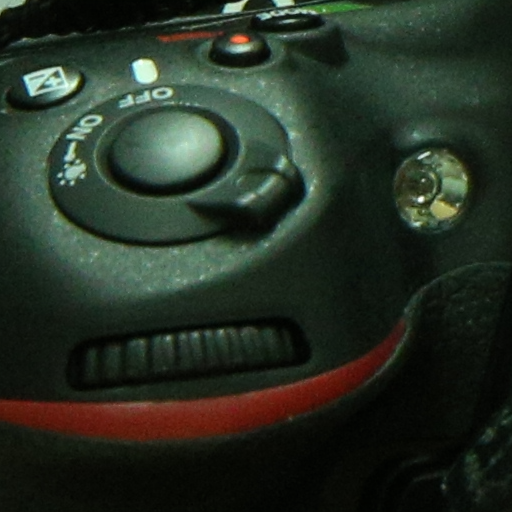}&
\includegraphics[width=.15\textwidth]{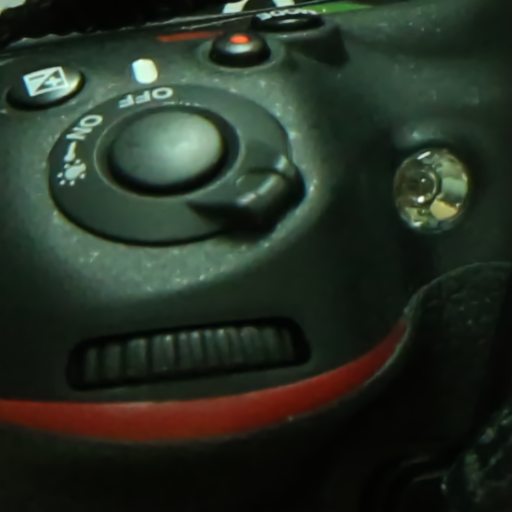}&
\includegraphics[width=.15\textwidth]{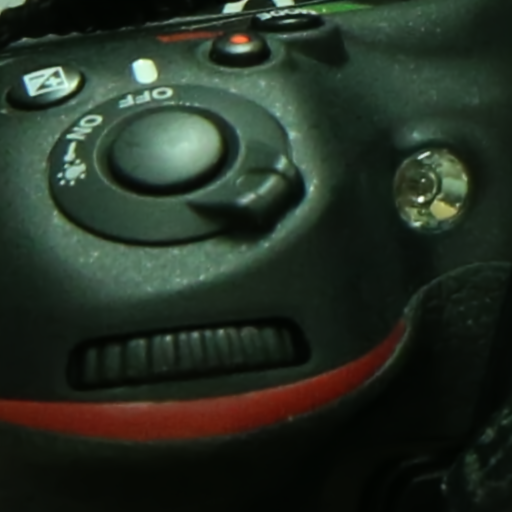}\\
DnCNN (37.56) & CBDNet (40.40) & Ours (40.50) \\
\end{tabular}
\end{center}
\vspace*{-2mm}
\caption{An image from Nam dataset~\cite{nam2016holistic} with JPEG compression. CBDNet is trained explicitly on JPEG compressed images; still, our method performed better.}
\label{fig:Nam}
\vspace*{-3mm}
\end{figure}

\begin{figure}
\begin{center}
\begin{tabular}[b]{c@{ }c@{ }c@{ }c} 
      
\includegraphics[width=.11\textwidth]{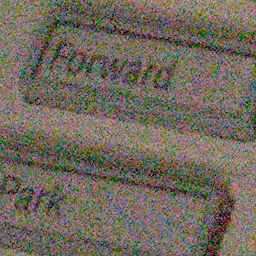}&   
\includegraphics[width=.11\textwidth]{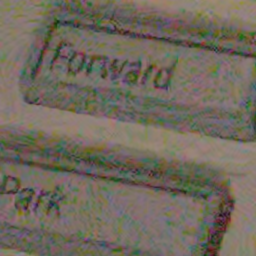}&
\includegraphics[width=.11\textwidth]{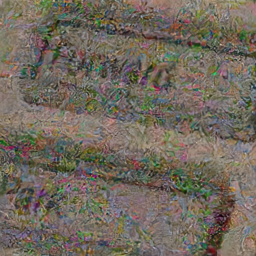}&
\includegraphics[width=.11\textwidth]{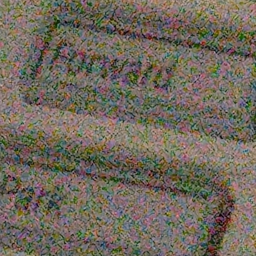}\\
 & 25.75 dB& 21.97 dB& 20.76 dB\\
Noisy & CBM3D & IRCNN  & DnCNN \\

\includegraphics[width=.11\textwidth]{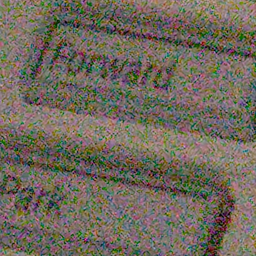}&
\includegraphics[width=.11\textwidth]{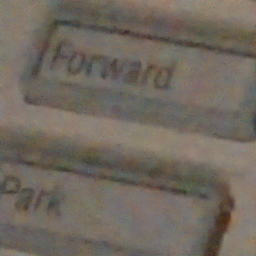}&
\includegraphics[width=.11\textwidth]{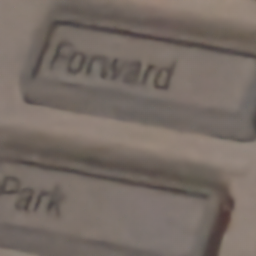}&
\includegraphics[width=.11\textwidth]{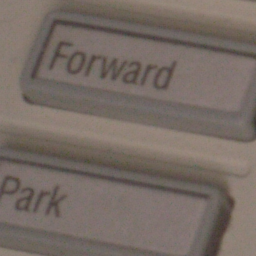}\\
19.70 dB & 28.84 dB& \textbf{35.57 dB}&  \\
FFDNet  & CBDNet  & Ours& GT \\

\end{tabular}
\end{center}
\vspace*{-2mm}
\caption{A challenging example from SSID dataset~\cite{abdelhamed2018high}. Our method can remove noise and restore true colors.}
\label{fig:SSID}
\vspace*{-3mm}
\end{figure}

\paragraph{SSID:} As a last dataset, we employ the SSID real noise dataset which has the highest number of test (validation) images available. The results in terms of PSNR are shown in the second row of Table~\ref{table:Nam_SSID}. Again, it is clear that our method outperforms FFDNet~\cite{zhang2018ffdnet} and CBDNet~\cite{guo2018CBDnet} by a margin of 9.5dB and 7.93dB, respectively.  In Figure~\ref{fig:SSID}, we show the denoised results of a challenging image by different algorithms. Our technique recovers the true colors which are closer to the original pixel values while competing methods are unable to restore original colors and in specific regions induce false colors.

\section{Conclusion}

In this paper, we present a new CNN denoising model for synthetic noise and real noisy photographs. Unlike previous algorithms, our model is a single-blind denoising network for real noisy images. We propose a novel restoration module to learn the features and to enhance the capability of the network further; we adopt feature attention to rescale the channel-wise features by taking into account the dependencies between the channels. We also use LSC, SSC, and SC to allow low-frequency information to bypass so the network can focus on residual learning.  Extensive experiments on three synthetic and four real-noise datasets demonstrate the effectiveness of our proposed model.\let\thefootnote\relax\footnotetext{This work was supported in part by NH\&MRC Project grant~\#~1082358.}

{\small
\bibliographystyle{ieee_fullname}
\bibliography{ref}
}

\end{document}